\pdfoutput=1

\documentclass[11pt]{article}

\usepackage[final]{acl}

\usepackage{times}
\usepackage{latexsym}

\usepackage[T1]{fontenc}

\usepackage[utf8]{inputenc}

\usepackage{microtype}

\usepackage{inconsolata}

\usepackage{graphicx}
\usepackage{colortbl}
\usepackage{geometry}
\usepackage{multirow}

\title{MemeMind at ArAIEval Shared Task: Spotting Persuasive Spans in Arabic Text with Persuasion Techniques Identification}

\author{
\textbf{Md Rafiul Biswas\textsuperscript{1}},
\textbf{Zubair Shah\textsuperscript{1}},
\textbf{Wajdi Zaghouani\textsuperscript{1,2}}
\\
\textsuperscript{1} Hamad Bin Khalifa University, Qatar Foundation, Doha, Qatar,
\\
\textsuperscript{2} Northwestern University in Qatar, Education City, Doha,
Qatar
  \\
  \small{
   \textbf{Correspondence:} \href{mailto:wajdi.zaghouani@northwestern.edu}{wajdi.zaghouani@northwestern.edu}
}}

\begin{document}
\maketitle
\begin{abstract}
This paper focuses on detecting propagandistic spans and persuasion techniques in Arabic text from tweets and news paragraphs. Each entry in the dataset contains a text sample and corresponding labels that indicate the start and end positions of propaganda techniques within the text. Tokens falling within a labeled span were assigned "B" (Begin) or "I" (Inside), "O", corresponding to the specific propaganda technique. Using attention masks, we created uniform lengths for each span and assigned BIO tags to each token based on the provided labels. Then, we used AraBERT-base pre-trained model for Arabic text tokenization and embeddings with a token classification layer to identify propaganda techniques. Our training process involves a two-phase fine-tuning approach. First, we train only the classification layer for a few epochs, followed by full model fine-tuning, updating all parameters. This methodology allows the model to adapt to the specific characteristics of the propaganda detection task while leveraging the knowledge captured by the pre-trained AraBERT model. Our approach achieved an F1 score of 0.2774, securing the 3rd position in the leaderboard of Task 1. 
\end{abstract}

\section{Introduction}

Propagandistic detection techniques involve identifying content that aims to influence an audience's perception or persuade them to a certain viewpoint using biased, misleading, or emotionally charged information \cite{martino2020survey}. This field of study is crucial due to the prevalence of propaganda in various forms of media, including news articles, social media posts, and advertisements. Techniques such as appeal to fear, loaded language, bandwagon, and testimonials are commonly used to manipulate public opinion. These techniques determine the presence of biased language and measure the emotional impact of the messages conveyed, thus providing a systematic way to sift through vast amounts of data and uphold the integrity of information consumed by the public. \cite{veeramani2023knowtellconvince,li2020syrapropa, khanday2021detecting, abdullah2022detecting}.

Recent studies have explored various aspects of Arabic natural language processing that are relevant to propaganda detection \cite{alam2024armeme}. \citealp{farha2021overview} organized a shared task on sarcasm and sentiment detection in Arabic, highlighting the importance of understanding the nuances of language use in social media. \citealp{charfi2019fine} developed a fine-grained annotated multi-dialectal Arabic corpus, which can be valuable for training models to handle the linguistic diversity of Arabic text.
The COVID-19 pandemic has also brought attention to the spread of disinformation and the need for effective detection methods. \citealp{alam2021fighting} modeled the perspectives of various stakeholders in fighting the COVID-19 infodemic, emphasizing the importance of a collaborative approach. \citealp{shurafa2020political}  analyzed political framing in the context of the pandemic, demonstrating how language can be used to shape public opinion. 

Shared Task 1 \cite{araieval:arabicnlp2024-overview} aimed to identify the propaganda techniques utilized within a diverse range of textual forms, such as news paragraphs or tweets, and to precisely locate the sections of text where each propaganda technique is employed. The shared task 1 is in Arabic dataset. While working with propaganda detection, finding textual spans is challenging, and it becomes more complicated regarding Arabic data. Arabic is a morphologically rich language with complex grammatical structures and a wide range of dialects \cite{rabiee2011adapting}. This complexity adds layers of difficulty to tasks like part-of-speech (POS) tagging and syntactic analysis, which are essential for identifying propaganda techniques \cite{dehdari2011morphological}. There is a lack of annotated datasets for propaganda detection in Arabic compared to languages like English. This scarcity makes it challenging to train and evaluate machine learning models effectively, as they heavily rely on annotated data for supervised learning.

In this paper, we propose an approach that leverages the AraBERT-base pre-trained model for Arabic text tokenization and text embeddings, combined with a token classification layer to detect propaganda techniques present in the text. We evaluate our model using the dataset provided by the ArAIEval Shared Task and compare our results with those of other participants.

\section{Related Work}
The growing prevalence of propaganda in digital media has necessitated the development of automated tools that can effectively detect and annotate propaganda techniques in text. Various approaches have been proposed to tackle this challenge, ranging from traditional machine learning methods to deep learning techniques. Early work in propaganda detection focused on identifying propagandistic news articles using machine learning classifiers. \citealt{barron2019proppy} developed a system called PropPy, which utilized a combination of stylistic, lexical, and syntactic features to detect propaganda at the article level. They achieved promising results, demonstrating the effectiveness of leveraging linguistic cues for propaganda identification.
With the advent of deep learning, researchers began exploring neural network architectures for propaganda detection. Bi-LSTM (Bidirectional Long Short-Term Memory) models have been widely used to capture sequential dependencies in text. Mapes et al. \cite{mapes2019divisive} employed a Bi-LSTM model with an attention mechanism to detect propaganda techniques in news articles. Their approach showcased the benefits of utilizing deep learning for capturing complex patterns in propagandistic text.

Recent studies have focused on fine-grained propaganda detection, aiming to identify specific propaganda techniques at the sentence or fragment level.\citealt{da2019fine} introduced a new dataset, named SemEval-2020 Task 11, which consists of news articles annotated with 18 propaganda techniques. They proposed a multi-granularity neural network that learns representations at different levels of text granularity, enabling the detection of propaganda techniques at both the sentence and fragment levels.
The ArAIEval Shared Tasks \cite{araieval:arabicnlp2024-overview, araieval:arabicnlp2023-overview} have focused on the detection of propagandistic techniques and disinformation in Arabic text, providing a platform for researchers to develop and evaluate their models. \citealt{hasanain2024can} investigated the potential of GPT-4, a large language model, in identifying propaganda spans in news articles. They also explored the use of large language models for propaganda span annotation \cite{hasanain2023large}.

Pre-trained language models, such as BERT \cite{devlin2018bert} and its variants, have revolutionized NLP tasks, including propaganda detection. These models capture rich linguistic knowledge from large-scale unsupervised pre-training, which can be fine-tuned for specific tasks. AraBERT \cite{antoun2020arabert} is a pre-trained language model specifically designed for Arabic text, which has shown promising results in various Arabic NLP tasks.
Our work extends prior research by utilizing the AraBERT-base pre-trained model for Arabic propaganda detection. We utilized a token classification method that merges AraBERT's embeddings with a classification layer to detect propaganda spans and techniques. Our dual-phase fine-tuning strategy enables the model to adapt to the nuances of propaganda detection while leveraging the pre-trained AraBERT model's knowledge. Overall, our contribution complements existing efforts in Arabic propaganda detection, aiming to enhance detection accuracy and mitigate manipulative content proliferation.

\section{Data}
\begin{figure}
    \centering
    \includegraphics[width=1.0\columnwidth]{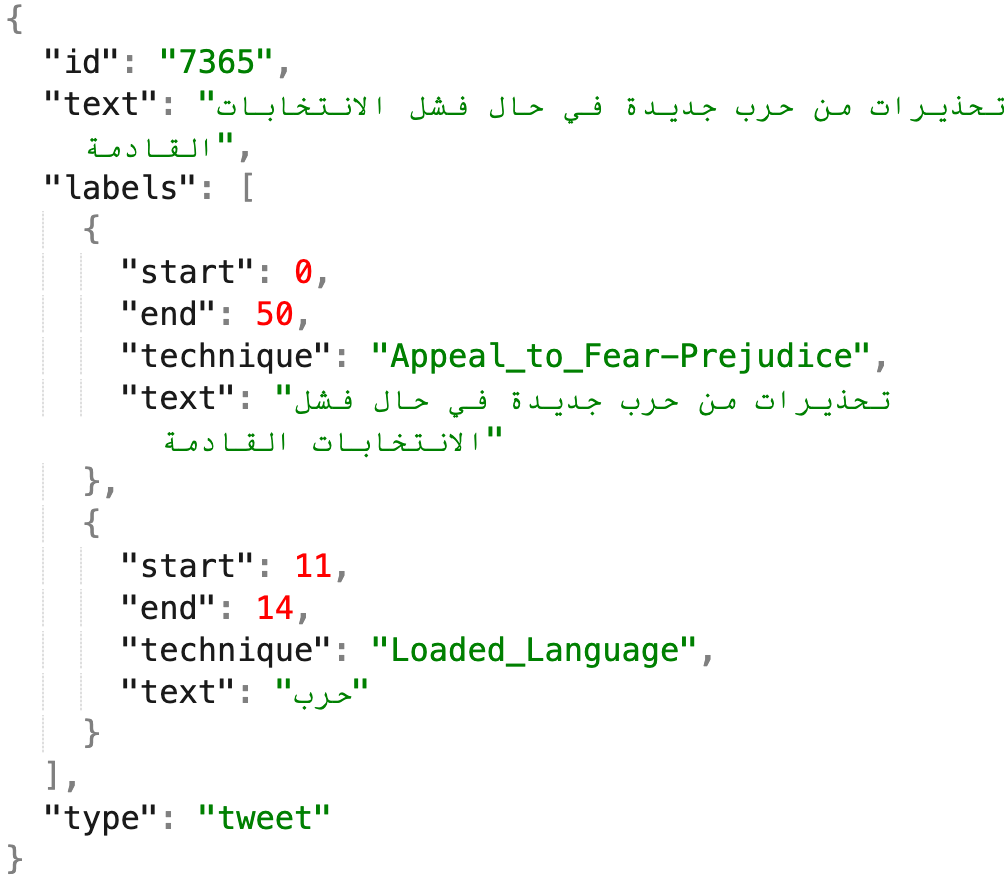}
    \caption{A Snippet of Data}
    \label{fig:jsonfile}
\end{figure}

The dataset used was provided by the organizers of the ArAIEval Shared Task 1 \cite{araieval:arabicnlp2024-overview}. The dataset comprises Arabic text snippets, including news paragraphs and tweets, annotated with propaganda spans and corresponding persuasion techniques. Figure \ref{fig:jsonfile} shows a snippet of the dataset. It contains a unique "id" identifier for each data. The "text" label is the key element to finding the propaganda span. Each data can contain either one label or multiple labels. There are a total of 23 techniques. Each technique has a start and end within the text. 

The training set comprises 6,997 text snippets, while the validation set contains 921 snippets. The test set used for the final evaluation consists of 1046 text snippets. The label distribution across the dataset is imbalanced, with some propaganda techniques being more prevalent than others. The most common techniques in the training set are "Loaded Language" (55.69\%), "Name Calling Labeling" (14.23\%), and "Exaggeration-Minimisation" (6.83\%).

Arabic is known for its diverse dialects, and the dataset contains a mix of Modern Standard Arabic and various regional dialects. This linguistic diversity adds complexity to the task of propaganda detection, as different dialects may express persuasion techniques differently. One of the challenges posed by the ArAIEval dataset is the presence of code-switching and dialectal variations in the text snippets. Preprocessing steps were applied to the dataset before feeding it into our system. We utilized the AraBERT tokenizer, specifically designed for Arabic text, to tokenize the input snippets. The tokenizer handles Arabic-specific characteristics, such as discretization and morphological segmentation.

\section{System}
\begin{figure}[htbp]
    \centering
    \includegraphics[width=1.0\columnwidth]{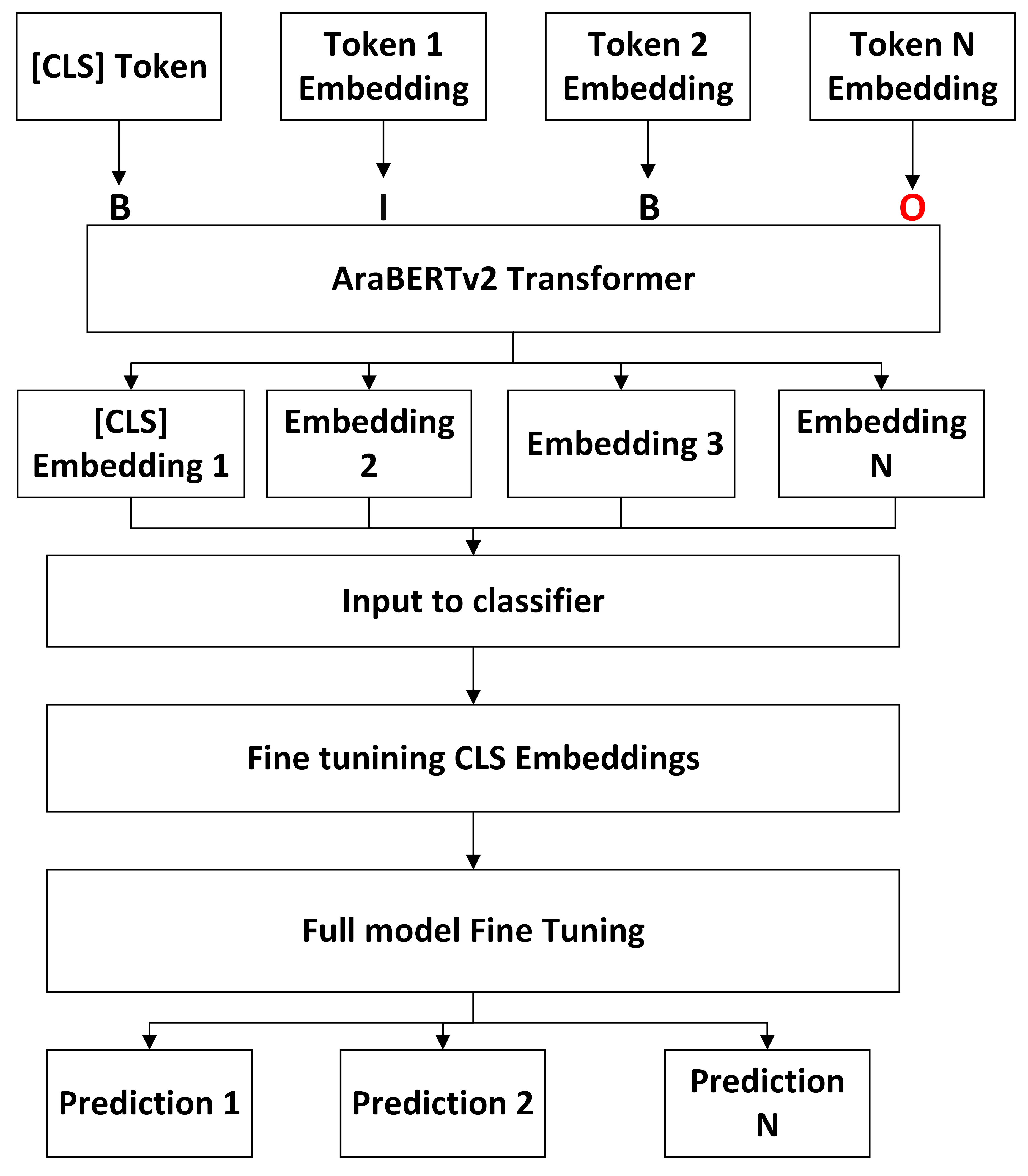}
    \caption{Flowchart for Task1}
    \label{fig:enter-label}
\end{figure}

Our proposed system architecture leverages the AraBERT-base pre-trained model, which is a transformer-based model specifically trained on a large corpus of Arabic text. The flowchart shows the process Fig.\ref{fig:enter-label}.
Firstly, we encoded each label into hard-coded and created a dictionary to hold the label to ID mapping. Then, we created bio tags for each entry. We utilized the AraBERT-base "AutoTokenizer" for text tokenization and embeddings, capitalizing on its ability to tokenize the text sequences, ensuring compatibility with pre-trained language models. For each text entry, the tokenizer generates tokenized representations and BIO (Beginning, Inside, Outside) tags indicating the presence of propaganda techniques at the token level. 
The tokenized input is passed through the AraBERT-base model to obtain contextualized word embeddings. These embeddings are then fed into a token classification layer consisting of a linear layer followed by a softmax activation function. The token classification layer predicts the propaganda technique associated with each token in the input sequence. We employ a two-phase fine-tuning approach to adapt the pre-trained AraBERT-base model to the task of propaganda detection. In the first phase, we freeze the parameters of the AraBERT-base model and train only the token classification layer for a few epochs. This allows the classification layer to learn the task-specific mapping from the pre-trained embeddings to the propaganda labels.
In the second phase, we unfreeze the parameters of the AraBERT-base model and fine-tune the entire model end-to-end. This enables the model to adapt its representations to the specific characteristics of the propaganda detection task.
The hyperparameters used in our system were determined through experimentation and validation. We used a batch size of 32, a learning rate of 2e-5 for the token classification layer, and a learning rate of 1e-5 for fine-tuning the AraBERT-base model. The model was trained for 10 epochs in the first phase and 5 epochs in the second phase.
The training was conducted on a single NVIDIA GeForce RTX 2080 Ti GPU, and the total training time was approximately 6 hours. The source code for our system is available at \url{https://github.com/rafiulbiswas/ArAIEval_ArabicNLP24_Task_1} for reproducibility.
\section{Results}
\subsection{Model Performance} Table \ref{tab:performance} shows different model performance for this dataset. In our evaluation of Arabic BERT-based models for a bio-tagging task, performance varied across key metrics. The bert-base-arabertv model achieved a moderate Micro-F1 score of 0.237 and a low Macro-F1 score of 0.020, indicating challenges in capturing relevant instances consistently. Despite decent recall (0.379), its precision was relatively lower (0.265), suggesting a tendency for false positives. In the case of the bert-base-arabertv2 model, the addition of hyperparameter tuning, such as attention-mask and adjustments using PyTorch, likely enabled the model to focus on relevant parts of the input sequence and optimize its learning process accordingly showed improved performance, with a Micro-F1 of 0.277 and a Macro-F1 of 0.058, balancing accuracy and recall better (0.245 and 0.320).

\title{Model Performance Comparison}
\author{}
\date{}
\maketitle
\begin{table}[!ht]  
\centering
\begin{tabular}{|>{\raggedright\arraybackslash}p{1.8cm}|>{\centering\arraybackslash}p{0.9cm}|>{\centering\arraybackslash}p{0.9cm}|>{\centering\arraybackslash}p{0.9cm}|>{\centering\arraybackslash}p{0.9cm}|}
\hline
\textbf{Model} & \textbf{Micro-F1} & \textbf{Macro-F1} & \textbf{Prec-ision} & \textbf{Recall} \\
\hline
AraBERT  & 0.237 & 0.020 & 0.265 & 0.379 \\
\hline
AraBERTv2  & 0.277 & 0.058 & 0.245 & 0.320 \\
\hline
mBERT  & 0.130 & 0.009 & 0.179 & 0.231 \\
\hline
CAMeLBERT-da  & 0.199 & 0.067 & 0.210 & 0.261 \\
\hline
\end{tabular}
\caption{Performance Metrics for Various Models}
\label{tab:performance}
\end{table}

In contrast, the UBC-NLP/MARBERT model underperformed with the lowest Micro-F1 (0.130) and Macro-F1 (0.009), reflecting challenges in precision and generalization. The CAMeL-Lab/bert-base-arabic-camelbert-da model achieved intermediate performance (Micro-F1: 0.199, Macro-F1: 0.067), suggesting room for improvement.
Overall, while each model had unique strengths and weaknesses, the bert-base-arabertv2 model showed the highest performance using bio-tagging tasks.

\subsection{Leaderboard} Our system achieved an F1 score of 0.2774 on the test set of the ArAIEval Shared Task 1, securing the 3rd position on the leaderboard. The F1 score is the harmonic mean of precision and recall, providing a balanced measure of the system's performance.
In addition to the F1 score, we also evaluated our system using other metrics, such as precision and recall. The precision of our system was 0.2446, indicating the proportion of correctly predicted propaganda spans among all the predicted spans. The recall was 0.3202, representing the proportion of actual propaganda spans that were correctly identified by our system. Table \ref{tab:leaderboard} shows the comparative analysis of our results with other participants in the leaderboard. We scored third in the competition. 
\begin{table}[ht]  
\centering
\begin{tabular}{|l|c|}
\hline
\rowcolor{lightgray} \textbf{Team} & \textbf{Micro F1} \\ \hline
CUET\_sstm & 0.2995 \\ \hline
Mela & 0.2833 \\ \hline
\rowcolor{green} Meme\_mind & 0.2774 \\ \hline
Nullpointer & 0.2541 \\ \hline
Sussex-AI & 0.1228 \\ \hline
SemanticCUETSync & 0.0783 \\ \hline
\rowcolor{pink} Baseline (Random) & 0.0151 \\ \hline
\end{tabular}
\caption{ArAIEval24 Task 1 Leaderboard}
\label{tab:leaderboard}
\end{table}
After the submission deadline, we experimented with some modifications to our system and obtained unofficial results. One notable improvement was the incorporation of additional linguistic features, such as part-of-speech tags and named entity recognition, which helped disambiguate similar propaganda techniques. These unofficial results suggest potential avenues for further enhancing the performance of our system.

\subsection{Error Analysis}
To gain insights into the performance of our system, we conducted an error analysis on the development set. We observed that our system struggled with certain types of propaganda techniques, particularly those that rely on subtle linguistic cues or require a deeper understanding of the context. For example, the technique of "Whataboutism" often involves drawing false equivalences or shifting the focus of the discussion, which can be challenging to detect based solely on the text.

Another common type of error was the misclassification of propaganda techniques that share similar characteristics. For instance, the techniques of "Loaded Language" and "Name Calling/Labeling" often involve the use of emotionally charged or pejorative terms, leading to confusion between the two categories.
Given more time and resources, we would also focus on conducting a more extensive error analysis and fine-grained evaluation of our system. This would involve analyzing the performance of specific propaganda techniques, identifying common error patterns, and developing targeted strategies to address them. 

\subsection{Limitations and Future Directions}
\textbf{Limitations:} The propaganda detection task in Arabic text poses several challenges, as evidenced by the performance of our system and the overall results of the ArAIEval Shared Task 1. One major challenge is the linguistic diversity of Arabic, with its various dialects and code-switching patterns. Dealing with this variability requires robust models that can handle different linguistic variations and capture the nuances specific to each dialect.

Another challenge is the subjectivity involved in identifying propaganda techniques. While some techniques, such as "Repetition" or "Slogans", are relatively straightforward to detect based on surface-level features, others, such as "Appeal to Fear/Prejudice" or "Bandwagon", require a deeper understanding of the context and the intended effect on the reader. This subjectivity makes it difficult to establish clear boundaries between propaganda and non-propaganda content.

The limited availability of annotated data for propaganda detection in Arabic is another significant challenge. The ArAIEval dataset provides a valuable resource, but the size of the training set is relatively small compared to datasets in other languages. This data scarcity limits the ability of models to learn the full spectrum of propaganda techniques and generalize well to unseen examples.

\noindent \textbf{Future Directions:} To address these challenges and improve the performance of our system, several strategies can be explored. One approach is to incorporate additional linguistic features and external knowledge sources to provide more contextual information to the model. This can include leveraging named entity recognition, sentiment analysis, or fact-checking datasets to help disambiguate between propaganda and non-propaganda content.
Incorporating additional linguistic features, exploring more advanced neural architectures, or leveraging larger pre-trained models specifically designed for propaganda detection.

Another direction is to explore more advanced neural architectures, such as graph neural networks or attention-based models, which can better capture the complex dependencies and long-range context in the text. 
Data augmentation techniques can also be employed to mitigate the issue of limited training data. Techniques such as back-translation, synonym replacement, or generating synthetic examples based on templates could help expand the training set and improve the model's robustness.

\section{Conclusion}
Our participation in the ArAIEval Shared Task 1 contributes to the ongoing efforts in propaganda detection and showcases the potential of leveraging pre-trained language models for this task in the Arabic language. Our proposed system, combining the Arabert-base model with token classification, achieved promising results and provides a foundation for further research and development in this critical area of NLP.
As the landscape of digital information continues to evolve, it is essential to continue advancing the state-of-the-art in propaganda detection. We hope that our work inspires further research and collaborations to tackle this important challenge, not only in Arabic but also in other languages and across various domains.

\section*{Acknowledgments}
The research work is partially funded by grant NPRP14C0916-210015 from the Qatar National Research Fund (QNRF) of the Qatar Research Development and Innovation Council (QRDI).

\bibliography{main}
\appendix
\section{Appendix}
\label{sec:appendix}

\subsection{Dataset Description}

\begin{figure}[!htbp]
    \centering
    \includegraphics[width=0.9\columnwidth]{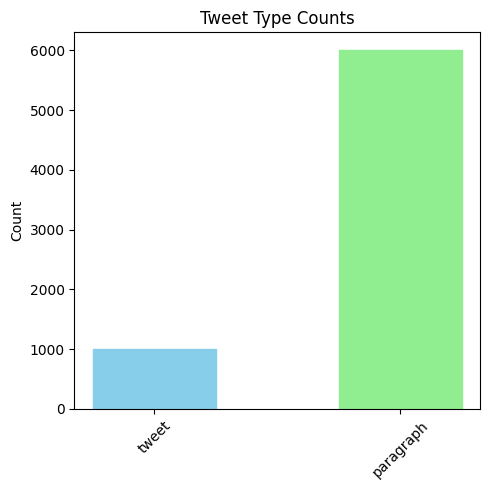}
    \caption{Training vs Test Dataset}
    \label{fig:tweet_type}
\end{figure}

Figure \ref{fig:tweet_type} shows two data types. There are tweet 995 (7.03\%) and paragraph 6002 (92.96\%).

\begin{figure}[!htbp]
    \centering
    \includegraphics[height=7cm,width=1.0\columnwidth]{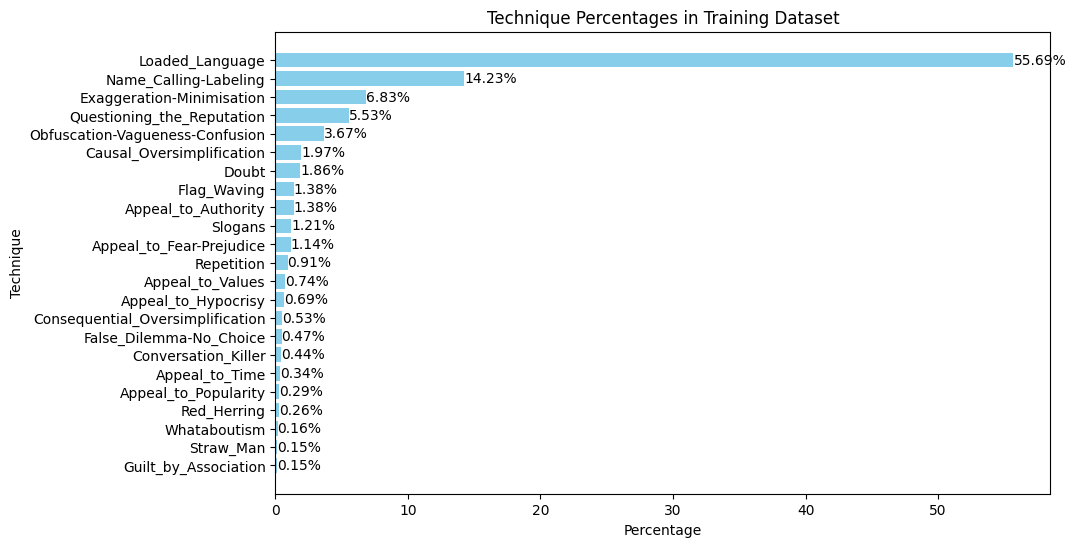}
    \caption{Techinque counts}
    \label{fig:techinque_count}
\end{figure}

Figure \ref{fig:techinque_count} depicts the distribution of rhetorical techniques found within a training dataset, highlighting the prevalence of various techniques used in language. The most prevalent technique is Loaded Language, accounting for a staggering 55.69\% of instances in the dataset. This indicates a significant use of emotive or biased terminology that aims to influence an audience's perception. Name Calling or Labeling follows, comprising 14.23\% of the dataset. This technique involves derogating a person or group to sway the audience's opinion. Other notable techniques include Exaggeration/Minimisation at 6.83\% and Questioning the Reputation at 5.53\%, which involve overstating or understating facts, and casting doubt on someone's character, respectively. Overall, the chart provides a clear visual representation of how different rhetorical techniques are employed in the dataset.

\subsection {Validation of Model}
\begin{figure}[!ht]
    \centering
    \includegraphics[width=0.9\columnwidth]{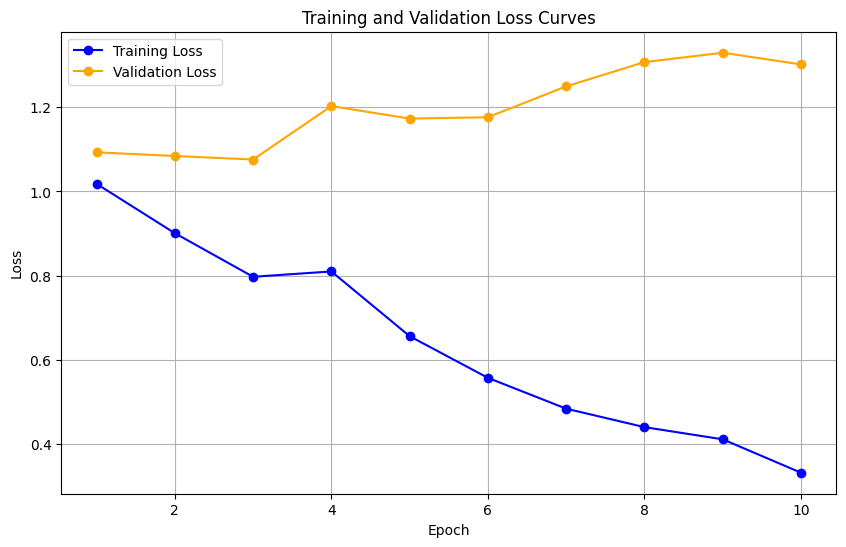}
    \caption{Loss between Training vs Validation Dataset}
    \label{fig:loss}
\end{figure}

Figure \ref{fig:loss} shows the model of the training and validation loss curves over 10 epochs. 
This is a good sign of decreasing training loss, indicating that the model is effectively learning from the training data. The increase in validation loss after a few epochs suggests overfitting. The model performs well on the training data but does not generalize well to the validation data.

Alternatively, the model is effectively learning from the training data. The decrease in validation accuracy after a few epochs suggests overfitting. The model performs well on the training data but does not generalize well to the validation data. 

\end{document}